\documentclass[letterpaper,journal]{IEEEtran}
\IEEEoverridecommandlockouts  
\usepackage{amsmath,amsfonts,amssymb}
\usepackage{array}
\usepackage{textcomp}
\usepackage{stfloats}
\usepackage{url}
\usepackage{verbatim}
\usepackage{graphicx}

\usepackage{gensymb,soul}
\usepackage{booktabs}
\usepackage[caption=false,font=footnotesize]{subfig}
\usepackage{epsfig} 
\usepackage{cite}
\usepackage{bm}
\usepackage[colorlinks,linkcolor=blue,citecolor=blue]{hyperref}
\usepackage{amssymb}
\usepackage{xcolor}
\usepackage{ifthen}
\usepackage{makecell}
\usepackage{algpseudocode}
\usepackage{threeparttable}
\usepackage[linesnumbered,ruled]{algorithm2e}

\DeclareMathAlphabet{\mathcal}{OMS}{cmsy}{m}{n}
\DeclareSymbolFont{largesymbols}{OMX}{cmex}{m}{n}
\usepackage{caption}
\usepackage{multirow}

\usepackage[utf8]{inputenc}
\usepackage{newunicodechar}
\newunicodechar{̌}{\v{}}

\definecolor{deepred}{RGB}{196, 49, 25}
\definecolor{lightblue}{rgb}{0.4, 0.6, 0.8}
\def\BibTeX{{\rm B\kern-.05em{\sc i\kern-.025em b}\kern-.08em
    T\kern-.1667em\lower.7ex\hbox{E}\kern-.125emX}}
\usepackage{balance}
\begin{document}

\title{LOVON: Legged Open-Vocabulary Object Navigator}

\author{Daojie Peng$^{1*}$, Jiahang Cao$^{1,2*}$, Qiang Zhang$^{1,2*}$, Jun Ma$^{1,3\dagger}$\\
$^1$The Hong Kong University of Science and Technology (Guangzhou)\\
$^2$Beijing Innovation Center of Humanoid Robotics\\
$^3$The Hong Kong University of Science and Technology\\
$^{*}$equal contributions; $^{\dagger}$corresponding author\\
\href{https://daojiepeng.github.io/LOVON/}{\color{lightblue}\textbf{{LOVON's Project Page}}\xspace}
}

\twocolumn[{
\renewcommand\twocolumn[1][]{#1}%
\maketitle
\vspace{-0.2in}
\begin{center}
    \centering
    \captionsetup{type=figure}
    \includegraphics[width=1\textwidth]{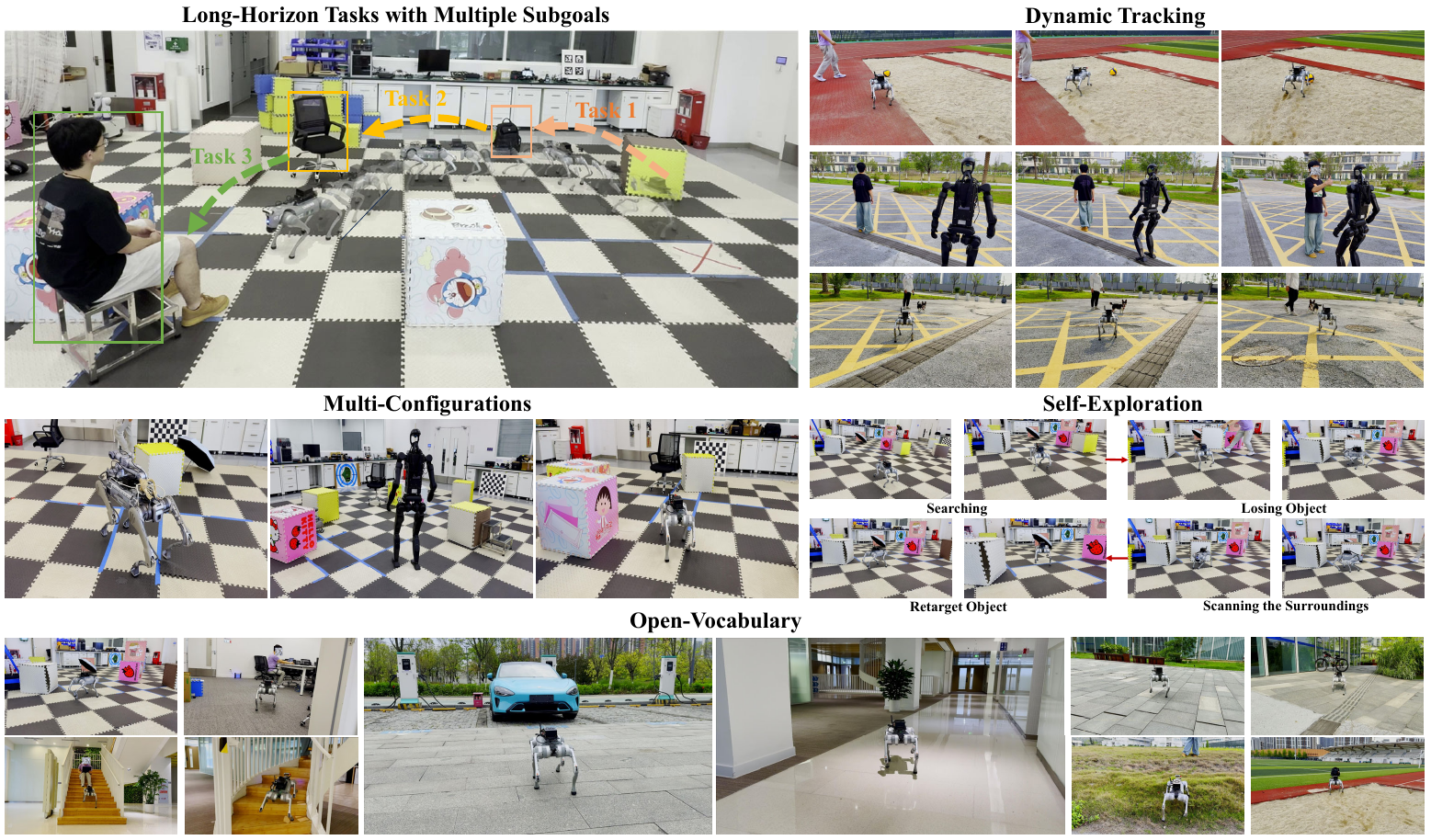}
    \caption{\textbf{Object navigation of legged robots in diverse open-world scenarios.} }
    \label{fig:teaser}
\end{center}
\vspace{-0.05in}
}]

\begin{abstract}
Object navigation in open-world environments remains a formidable and pervasive challenge for robotic systems, particularly when it comes to executing long-horizon tasks that require both open-world object detection and high-level task planning. 
Traditional methods often struggle to integrate these components effectively, and this limits their capability to deal with complex, long-range navigation missions. 
In this paper, we propose LOVON, a novel framework that integrates large language models (LLMs) for hierarchical task planning with open-vocabulary visual detection models, tailored for effective long-range object navigation in dynamic, unstructured environments.
To tackle real-world challenges including visual jittering, blind zones, and temporary target loss, we design dedicated solutions such as Laplacian Variance Filtering for visual stabilization. 
We also develop a functional execution logic for the robot that guarantees LOVON's capabilities in autonomous navigation, task adaptation, and robust task completion.
Extensive evaluations demonstrate the successful completion of long-sequence tasks involving real-time detection, search, and navigation toward open-vocabulary dynamic targets. 
Furthermore, real-world experiments across different legged robots (Unitree Go2, B2, and H1-2) showcase the compatibility and appealing plug-and-play feature of LOVON. 
\end{abstract}

\section{Introduction}
In recent years, large language models (LLMs)~\cite{achiam2023gpt} and vision models~\cite{jiang2022review, redmon2016you, caron2021emerging, zhang2022dino} have achieved revolutionary breakthroughs in the field of artificial intelligence. LLMs have significantly improved their capabilities in understanding and planning long-horizon tasks, enabling them to deeply comprehend complex contexts and generate efficient execution strategies, which brings new possibilities to task planning. Meanwhile, advances in open-vocabulary visual detection have empowered vision models to recognize and understand a diverse range of objects beyond predefined categories, greatly enhancing the adaptability of machine vision systems in scene understanding and object recognition. These leaps in perception and cognition lay a solid foundation for addressing complex long-horizon tasks in robotics.

Legged robots have evolved over decades and now demonstrate outstanding mobility in complex terrains. Their unique structural design and motion control allow them to adapt to various rugged environments, exhibiting terrain adaptability far beyond that of traditional wheeled robots. However, most current research focuses on optimizing single tasks, such as walking, jumping, climbing, and short-range navigation, lacking comprehensive consideration of complex long-horizon missions. It is acknowledged that the potential of legged robots to perform long-horizon tasks in open environments has not been fully explored, and that integrating advanced language and vision understanding with legged robot mobility is a key breakthrough for real-world applications.

In this paper, we propose LOVON, an innovative operating system that integrates the task planning capabilities of LLMs, the perception abilities of open-vocabulary visual detection, and a language-to-motion model (L2MM) for precise motion prediction, as shown in Fig.~\ref{fig:teaser}. The LOVON system addresses real-world challenges such as visual jitter caused by robot motion.
In particular, we design a Laplacian Variance Filtering technique to effectively mitigate visual instability during robot movement, ensuring the accuracy and continuity of object detection. We also propose the functional execution logic for robust task completion. Experiments on simulation benchmark Gym-Unreal~\cite{gymunrealcv2017} demonstrate the superior results of LOVON. In addition, we conduct extensive experiments on multiple legged platforms (Unitree Go2, B2, and H1-2), successfully accomplishing long-horizon tasks involving real-time detection, search, and navigation toward open-vocabulary dynamic targets. To the best of our knowledge, LOVON is the first operational system to achieve such complex capabilities in unstructured environments.
The main contributions of this work are summarized as follows:

\begin{itemize}
\item We propose LOVON, a unified framework that integrates LLMs, open-vocabulary visual detection, and a language-to-motion model, enabling the planning and execution of complex open-world long-horizon navigation tasks.
\item We develop a Laplacian Variance Filtering method to resolve dynamic blurring issues and improve the robustness of the system. Also, we introduce a robot execution logic that ensures adaptability to various environments.
\item We conduct comprehensive validation through simulations and real-world experiments across a variety of legged robot platforms. The results demonstrate that our system successfully performs open-vocabulary object search and navigation tasks in unstructured environments.
\end{itemize}

\section{Related Works}
\subsection{Large Language Models for Robotic Task Planning}
LLMs, such as GPT~\cite{achiam2023gpt}, 
and LLaMA~\cite{grattafiori2024llama}, have demonstrated remarkable capabilities in natural language understanding, reasoning, and task decomposition. In the context of robotics, LLMs have been increasingly adopted for high-level task planning, instruction following, and semantic reasoning. 
For example, SayCan~\cite{ahn2022can} integrates LLMs with robotic affordance models to map language instructions to executable actions, while Code as Policies~\cite{liang2023code} utilizes LLMs to generate code for robot controllers directly from natural language descriptions. Despite these advances, challenges remain in grounding language to real-world robot actions~\cite{fu2024humanplus, li2025amo, zhang2024whole}, handling ambiguous or user-specified instructions, and ensuring robust performance in unstructured environments. Our work builds upon these foundations by integrating LLM-based planning with open-vocabulary perception and legged robot mobility, aiming to address the limitations of previous approaches in long-horizon, open-world scenarios.

\subsection{Open-Vocabulary Visual Perception}

Open-vocabulary visual perception has evolved significantly from early fixed-class object detectors like Faster R-CNN~\cite{ren2016faster} and YOLO~\cite{redmon2016you}, which are confined to recognizing predefined categories and struggled in open-world scenarios. 
Subsequent works like Grounding DINO~\cite{liu2024grounding} have further enhanced the detection capability by integrating grounded pretraining to improve open-set detection accuracy
For robotic applications, real-time performance and robustness to dynamic camera motions, such as jittering caused by legged robot locomotion or temporary target occlusions, remain critical challenges~\cite{zhang2025parkour}. Existing methods often fail to maintain detection stability in such scenarios, lacking the feedback mechanisms to adapt to the robot’s motion state or environmental changes. LOVON addresses these gaps by developing specialized preprocessing techniques, like Laplacian variance filtering, to mitigate motion blur and ensure consistent visual input, while tightly integrating open-vocabulary detection with task planning and motion control for end-to-end execution in unstructured environments.

\subsection{Legged Robot Navigation and Long-Horizon Autonomy}

Legged robot navigation has advanced from low-level mobility to complex tasks, but existing approaches often focus on single-task optimization and lack integration of high-level planning for long-horizon missions~\cite{cheng2024navila, wang2025trackvla, sun2025trinity}. While legged robots show superior terrain adaptability~\cite{zhang2025distillation}, their potential for executing sequential goals in unstructured environments remains underexplored, as traditional systems often separate perception from motion planning and struggle with dynamic target tracking.  
LOVON addresses this by unifying hierarchical task decomposition with real-time motion control. An LLM-based planner breaks down long-horizon tasks, while the L2MM maps instructions and visual feedback to dynamic motion vectors. This allows adaptive behaviors like switching to search states when targets are lost, validated across Unitree Go2, B2, and H1-2 platforms in diverse terrains. By integrating open-vocabulary perception with legged mobility, LOVON enables autonomous robots to navigate complex environments and adapt to dynamic missions.

\begin{figure*}[t]
    \centering
    \includegraphics[width=1\textwidth]{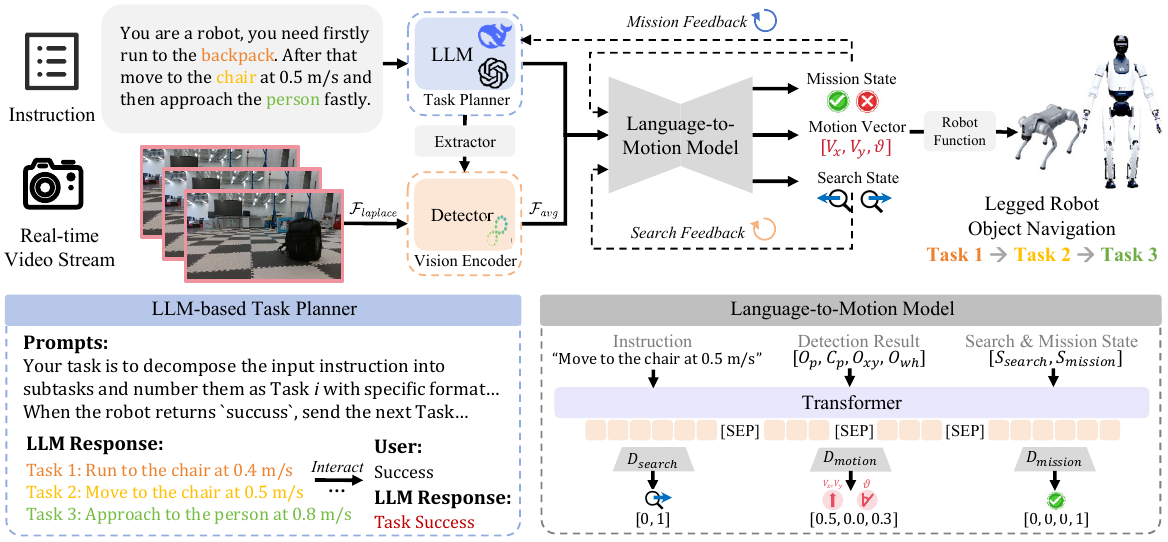}
    \caption{\textbf{Overview of LOVON's pipeline.} 
First, the LLM task planner reconfigures the human’s task into basic instructions, while the detection model processes the video stream using a Laplace filter. Then, the mission instructions, target object, bounding box, and states are input to the Language-to-Motion Model, which generates the robot's control vector and feedback, progressively completing all tasks.}
    \label{fig:vlm_model_pipeline}
    \vspace{-3mm}
\end{figure*}

\section{Problem Formulation}
\noindent\textbf{Task.} The task involves operating in an arbitrary open-world environment, where the robot is required to perform long-horizon tasks to search for different targets. The long-horizon task $T_{l}$, is defined as a set of subtasks $T_{l}=\{T_i|T_1, T_2, ...\}$, where each subtask corresponds to searching for a specific target $O_i$. The description of $T_{l}$ is flexible (example in the bottom left of Fig.~\ref{fig:vlm_model_pipeline}), allowing for varying mission objectives. The core challenge is for the robot to autonomously search for and identify different subgoals (targets), navigating toward them at different velocities based on mission instructions. These subgoals can vary throughout the task, requiring the robot to adapt dynamically.

\noindent\textbf{Goal.} Our goal is to develop a dual-system model: (I) A high-level policy that can decompose complex task instructions $T_{l}$ into individual subtasks with concrete instruction $I_\text{ins} = \{I_i |I_{1}, I_{2}, ...\}$ and perform task planning; (II) A low-level policy that, based on specific subtask instructions $I_i$ and video stream input $I_\text{RGB}$, can generate motion vectors $V_m \in R^3$ to achieve precise motion control. The model should be adaptable across various legged robots, ensuring versatility in real-world applications.

\section{Methodology}

\subsection{Overview}
The pipeline of LOVON is illustrated in Fig.~\ref{fig:vlm_model_pipeline}. Initially, the LLM reconfigures the human's long-horizon task into basic mission instructions. These instructions are then passed to an instruction object extractor (IOE) to identify the target object. The detection model processes the captured video stream, with the input image preprocessed using a Laplace filter. Finally, the mission instruction, target object, bounding box, mission state, and search state are combined as inputs to the proposed L2MM, which generates the robot's control vector and feedback states for both the LLM and L2MM.

\subsection{Multimodal Input Processing}
LOVON integrates two pretrained models: object detection model (e.g., \cite{yolo11_ultralytics, carion2020end, zhang2022dino, liu2024grounding}) for visual input processing and LLM (e.g., \cite{guo2025deepseek, achiam2023gpt, touvron2023llama}) for long-horizon task management. 
The input to the LLM consists of the system description $I_{sys}$, the user’s long-sequence task description $T_l$, and feedback from the L2MM $O_f$. Using this input, the LLM generates specific mission instructions ${I_i}$,
enabling LOVON to execute long-sequence tasks by producing the necessary instructions to achieve the mission objectives:
\begin{equation}
I_\text{ins} = f_\text{LLM}(I_{sys}, T_l, O_f).
\end{equation}
Then, our proposed IOE maps the instruction to the detection class. IOE uses a two-layer transformer with a perception layer to predict the object class:
\begin{equation}
I_\text{object} = f_\text{IOE}(I_m) \in \mathbf{C},
\end{equation}
where $\mathbf{C}$ represents the set of classes that the detection model is capable of recognizing.

Regarding the visual processing, the object detection model takes an RGB image $I_\text{RGB}$ and $I_\text{object}$ as input and outputs the desired detection information as follows:
\begin{equation}
    O_m, C_p, O_{xy}, O_{wh} = f_\text{det}(I_\text{RGB}, I_\text{object}).
\end{equation}
We use the normalized format for the detection results, with the predicted object denoted as $O_m$, the confidence score as $C_p$, and the center position of the bounding box as $O_{xy}=[x_n, y_n]$. The width and height of the bounding box are represented as $O_{wh}=[w_n, h_n]$. 
Additionally, we apply a moving average filter to smooth the bounding boxes from the object detection model’s output, further improving stability.

\begin{figure}[t]
    \centering
    \includegraphics[width=1\linewidth]{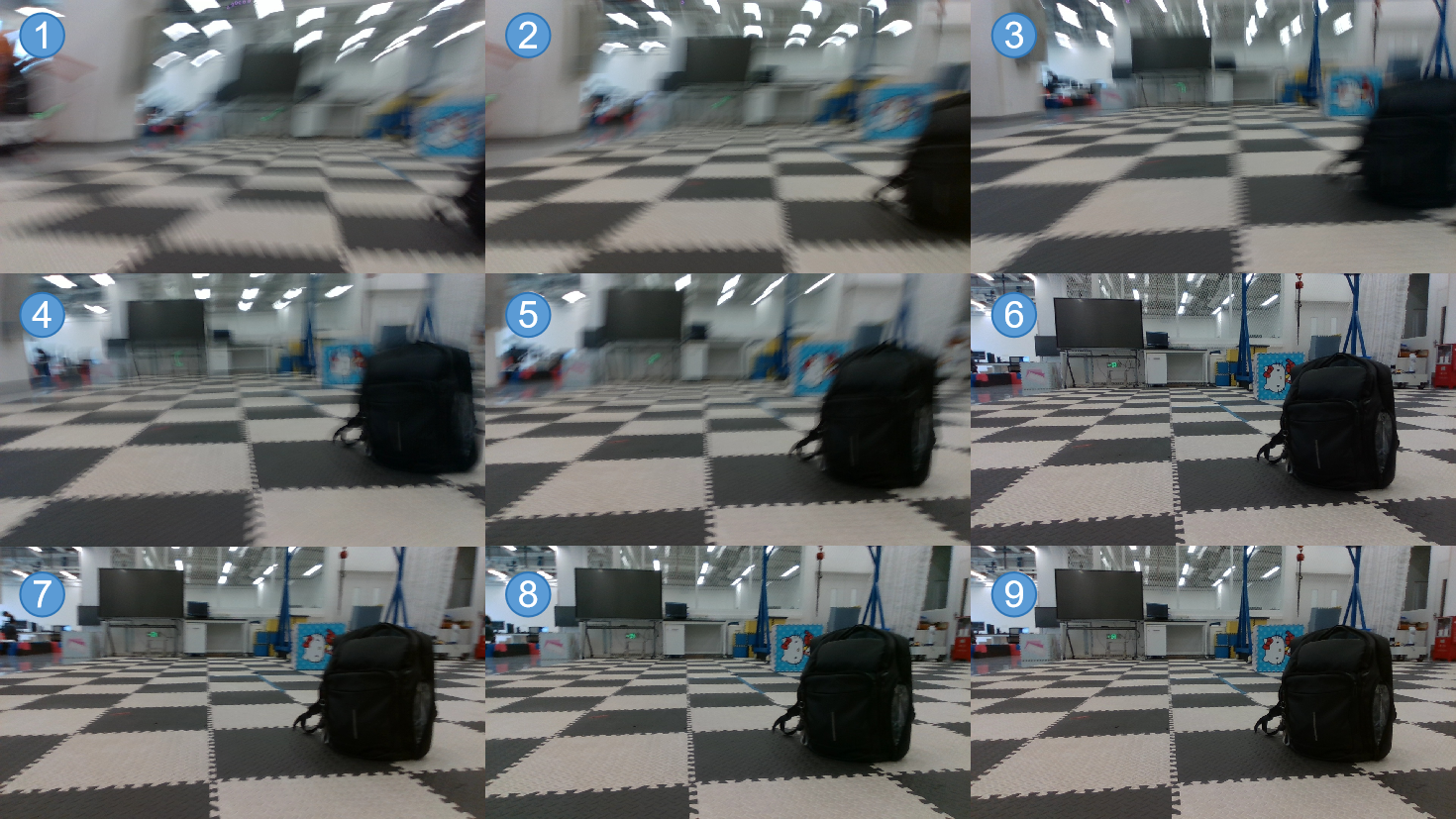}
    \caption{\textbf{Image blurring phenomenon.} This figure shows the occurrence of image blurring in the robot view, which impacts the clarity and accuracy of the processed images.}
    \label{fig:blur_views}
    \vspace{-3mm}
\end{figure}

\subsection{Laplacian Variance-Based Motion Blur Filtering}\label{subsec:motion-blur-method}

When the legged robot is in motion, the resulting fluctuations can cause motion blur in the captured frames, as shown in Fig.~\ref{fig:blur_views}. The first few frames are particularly blurred due to the robot’s dynamic locomotion, making them challenging for the vision model.
To address this, we propose a Laplacian variance-based method for detecting and filtering motion-blurred frames. This preprocessing step improves the robustness of inputs to the object-detection-based vision-language pipeline by mitigating the effects of motion blur and distortion caused by the robot’s movement and vibrations.

In particular, we first convert the RGB frame $I_\text{RGB}$ to grayscale $I_\text{gray}$. We then apply the Laplacian operator to enhance high-frequency components, yielding the Laplacian response. The variance of the Laplacian response is computed to assess the clarity of the frame. If the variance is below a threshold $T_{blur}$, the frame is classified as blurred and replaced by the last clear frame. The threshold $T_{blur}$ is empirically calibrated for robot scenarios. 
The performance of this filtering method is discussed in Sec. \ref{section:frame_filter}.

\subsection{Language-to-Motion Model}\label{sec:lmm}
Our proposed L2MM is the core module responsible for predicting motion and providing feedback.
L2MM is designed using an encoder-decoder architecture. The encoder takes a sequence of inputs which consists of the following components: the previous mission instruction \(I_{m0}\), the current mission instruction \(I_{m1}\), the predicted object \(O_p\), the predicted confidence \(C_p\), the center position \(O_{xy}\), the width and height of the normalized bounding box \(O_{wh}\), the current mission state \(S_{m}\), and the current search state \(S_{s}\). These inputs are concatenated, with each separated by special tokens [SEP], as
$I_{encoder} = \{ I_{m0}, I_{m1}, O_p,  C_p, O_{xy}, O_{wh}, S_{m}, S_{s} \}$
The encoder processes this sequence and outputs a latent state \(l_e\), which is then passed to the decoder 
that consists of three separate heads, each designed to handle different prediction tasks:

\noindent\textbf{Motion Vector Head} \(D_{motion}\). This head predicts the robot’s motion vector \(V_m\) based on the latent state \(l_e\). It is formulated as a sequence-to-vector problem, where the output is the control vector for the robot's movement:
\begin{equation}
    V_m = D_{motion}(l_e).
\end{equation}
    
\noindent\textbf{Mission State Head} \(D_{mission}\). This head predicts the mission state \(S_m\), which is used as feedback for the encoder to adjust future actions. The prediction is formulated as a sequence-to-number problem, where the output represents the current status of the mission. The prediction is given by:
\begin{equation}
    S_m = D_{mission}(l_e).
\end{equation}
    
\noindent\textbf{Search State Head} \(D_{search}\). This head predicts the search state \(S_s\), which indicates the robot's progress in searching for the target. It is also a sequence-to-number problem, where the output reflects the current state of the search:
\begin{equation}
    S_s = D_{search}(l_e).
\end{equation}

Each of these decoder heads uses a perception layer to process the latent state \(l_e\) and generate the respective outputs.

The final output of the model is a combination of the predictions from all three decoder heads:
\begin{equation}
O_{decoders} = \{ V_m, S_{m}, S_{s} \},
\end{equation}
where $V_m = [v_x, v_y, \theta]$, $v_x$ and $v_y$ represent the robot's velocities along the $x$ and $y$ axes, and $\theta$ represents the angular velocity; $S_{m}$ and $S_{s}$ is the output search state and mission state, respectively. The mission state includes \texttt{success} (task completed successfully) and \texttt{running} (moving towards the detected object). The search state includes \texttt{searching\_0} (searching by rotating left) and \texttt{searching\_1} (searching by rotating right). The relationship between states and their corresponding motion vectors is summarized in Table~\ref{tab:state_motion_mapping}.

This architecture enables the model to predict motion vectors, task states, and search states simultaneously, allowing the robot to not only control its motion accurately but also understand long task sequences and provide relevant feedback.

\begin{table}[t]
    \renewcommand{\arraystretch}{1.1}
    \centering
    \caption{\textbf{Relationship between states and motion vector.} In the running state, $v_x$ represents the moving speed of $x$ axis. The rotation $\theta_{corr}$ is regulated during execution to correct the heading direction.}
    \label{tab:state_motion_mapping}
    \begin{tabular}{c|l}
        \hline\hline
        State $S_m/S_s$ & Motion vector $V_m$ \\
        \hline
        success & $[0,\quad 0,\quad 0]$\\
        running & $[v_x,\ 0,\quad \theta_{corr}] $\\
        searching\_0 & $[0,\quad 0,\quad -0.3]$\\
        searching\_1 & $[0,\quad 0,\quad 0.3]$\\
        \hline\hline
    \end{tabular}
    \vspace{-3mm}
\end{table}

\subsection{Loss Functions}

The model is trained using different loss functions depending on the task:

\noindent\textbf{Motion Vector Loss}. For the motion vector head \(D_{motion}\), we use the mean squared error loss with coefficient $\beta$ to measure the difference between the predicted and actual motion vectors:
\begin{equation}
    L_{MSE} = \frac{1}{N} \beta\sum_{i=1}^{N} (V_{m_{pred}}^i - V_{m_{true}}^i)^2.
\end{equation}

\noindent\textbf{Mission and Search State Loss}. For the mission and search state heads \(D_{mission}\) and \(D_{search}\), we use cross-entropy loss to compare the predicted states with the ground truth labels:
\begin{equation}
    L_{CE} = - \sum_{i=1}^{N} y_i \log(p_i),
\end{equation}
where \(y_i\) is the true label, and \(p_i\) is the predicted probability for each class.

\begin{figure}[!t]
    \centering
    \includegraphics[width=1.0\linewidth]{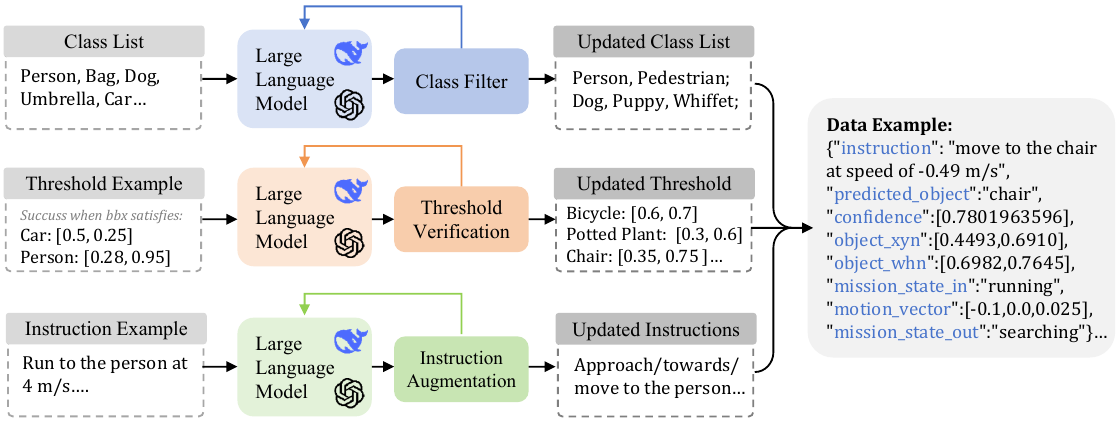}
    \caption{\textbf{Dataset generation pipeline.}
The pipeline includes three modules: expanding object class synonyms, generating instruction variations, and adapting detection thresholds for different object categories.}
    \label{fig:data_generation}
    \vspace{-3mm}
\end{figure}

\section{Robot Functional Logic for Task Execution}\label{sec:robot_functions}

Once the model generates predictions, the robot follows its functional logic to execute tasks and adapt to environmental changes. The key functions that guide its behavior during task execution include:

\begin{itemize}
    \item \textbf{Execute New Mission}: The robot compares the previous mission instruction with the current one. If they differ, the robot begins the new task.
    
    \item \textbf{Run to the Object}: Upon detecting the mission object, the robot navigates towards it based on the motion vector and the detection results.
    
    \item \textbf{Search for Lost Object}: If the robot loses track of the mission object, it automatically switches to a searching state and adjusts its motion to relocate the object.
    
    \item \textbf{Maintain the State}: The robot maintains its current state based on real-time visual inputs until a transition is triggered, ensuring consistent task execution.
    
    \item \textbf{Accomplish the Mission}: The robot monitors the mission object and, once $O_{wh}$ is within a success threshold, it stops and transitions to the \texttt{success} state.
\end{itemize}

These functional rules ensure autonomous navigation, task adaptation, and robust task completion.

\section{Dataset Preparation}\label{sec:dataset_preparation}

As illustrated in Fig.~\ref{fig:data_generation}, the dataset generation pipeline consists of three main components:

\noindent\textbf{Detection Class Synonym Expansion}.
We use an LLM to generate synonyms for the predefined object classes, enriching the object categories and improving the model’s ability to generalize across different object descriptions.

\noindent\textbf{Instruction Variation}.
To enhance the language module, we use the LLM to generate paraphrases of mission instructions. This allows the model to process diverse sentence structures while preserving core information, improving its adaptability.

\noindent\textbf{Threshold Generation for Object Categories}.
We define success thresholds for object detection based on initial examples, then use the LLM to adapt these thresholds for other categories, ensuring the model handles different object sizes. 

During the generation process, the generated data is fed back into the LLM to refine the dataset iteratively, avoiding redundancy and improving the dataset's diversity over time.

The dataset generation process is fast and easy to expand. It takes less than 15 minutes to generate 1 million data with CPU Intel i9-12900KF.

\begin{figure}[t]
    \centering
    \includegraphics[width=1.0\linewidth]{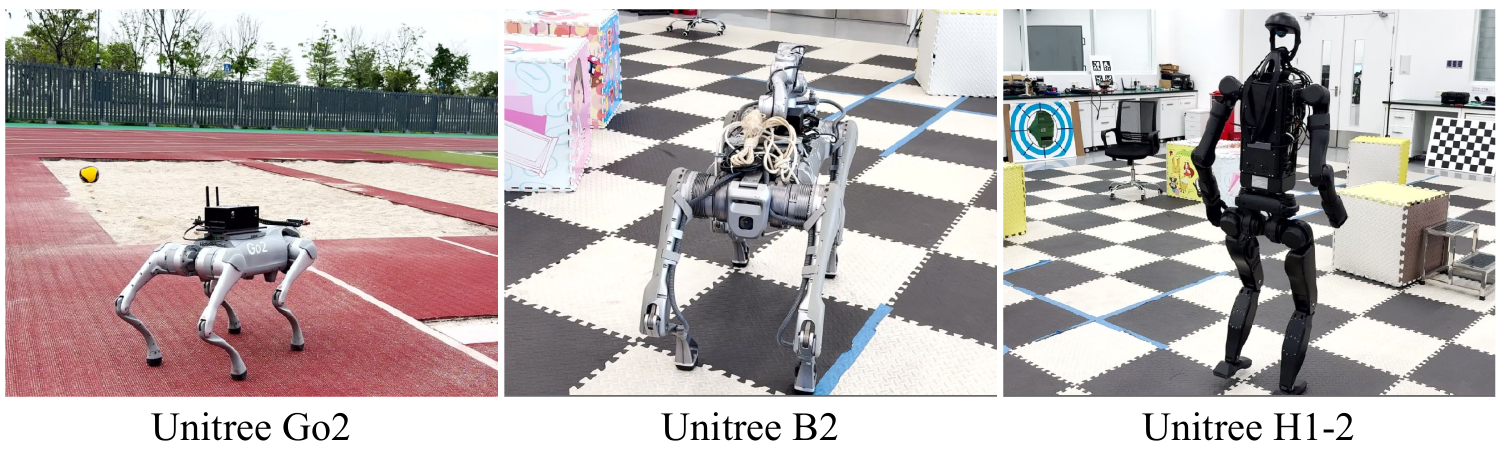}
    \caption{\textbf{LOVON with multi-configurations}. LOVON can be seamlessly adapted to any legged robot for precise object navigation.}
    \label{fig:multi_robots}
    \vspace{-3mm}
\end{figure}

\section{Experiment}

\subsection{Experiment Setup}
\noindent\textbf{Model Details.}
For object detection, we employ the recently developed and efficient YOLO-11~\cite{yolo11_ultralytics}, which offers both high performance and lightweight architecture. As the task planner and data generation assistant, we utilize DeepSeek R1~\cite{guo2025deepseek}.
L2MM is a transformer-based model 
with feature dimension 256, 4 layers, and 8 attention heads, a feedforward dimension of 1024, and a linear head layer. IOE shares the same general architecture, with a reduced feature dimension size of 64, 2 transformer layers with 4 attention heads, and a feedforward dimension of 256.

\noindent\textbf{Training Settings.}
Our collected dataset consists of 1 million samples, which are divided into training and testing sets in a 4:1 ratio. We use the NVIDIA RTX 3080 Ti GPU for training.
The L2MM model is trained with a dropout rate of 0.1, a learning rate of 
10$^{-4}$, a batch size of 512, a maximum sequence length of 64, and a motion loss coefficient $\beta$ set to 10. It is trained 
25 epochs using the AdamW optimizer. The total training time is approximately 1 hour.
Similarly, the IOE model is trained with the same training settings by approximately 30 minutes.

\noindent\textbf{Robot Settings.}
LOVON is versatile and can be applied to various legged robots. In our experiments, we evaluate three representative models: Unitree Go2, B2, Unitree H1-2, as shown in Fig.~\ref{fig:multi_robots}. For the computing platform, we utilize the Jetson Orin, while the visual platform consists of the robots' built-in cameras and the Realsense D435i camera.

\subsection{Performance of Motion-Blurred Frame Filtering \label{section:frame_filter}}

\begin{figure}[t]
    \centering
    \includegraphics[width=.9\linewidth]{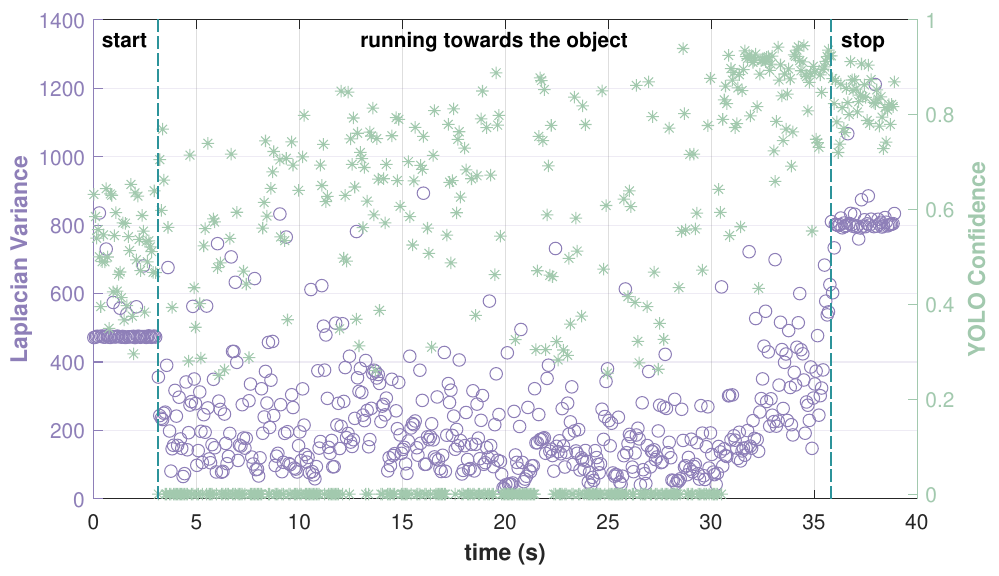}
    \caption{\textbf{Visualization of the relationship between Laplacian variance and the confidence of the detection object.} 
    }
    \label{fig:time_lap_conf_chair03_example}
    \vspace{-3mm}
    
\end{figure}

To investigate the impact of motion blur on object detection performance, we conduct experiments to analyze the relationship between Laplacian variance and detection confidence. We command the robot to approach a backpack, chair, or person at fixed speeds of 0.3, 0.5, or 0.7 m/s. The camera updates at approximately 15 Hz, ensuring the target remains in view. After capturing the frames, we compute the Laplacian variance for each frame and input it into the object detection model to obtain predicted confidence scores.
As shown in Fig.~\ref{fig:time_lap_conf_chair03_example}, the relationship between Laplacian variance and YOLO confidence fluctuates significantly, e.g., in the running phase, the detection model always fails to recognize the target object in many frames, despite the object remaining within view.

\begin{figure}[t]
    \centering
    \includegraphics[width=.85\linewidth]{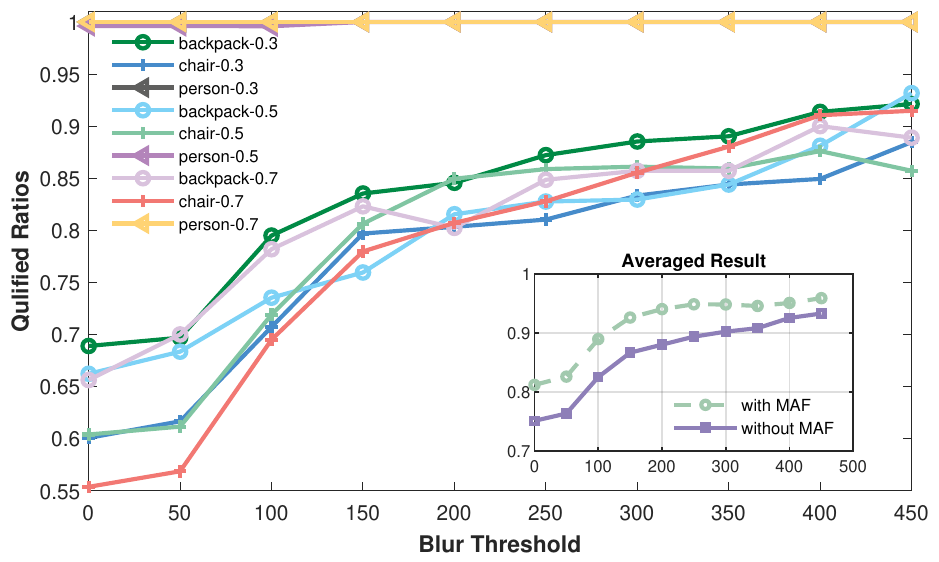}
    \caption{\textbf{Thresholds and qualified ratios of different objects at different speeds.} As shown in the subplot, our filter method with MAF increases the qualified frame rate by 25\%.
    }
    \label{fig:threshold_ratio_compare}
    \vspace{-3mm}
\end{figure}

To address this, we adopt our proposed motion-blurred frame filtering method. By setting a blur threshold introduced in Sec.~\ref{subsec:motion-blur-method}, we can filter out frames with excessive motion blur. The impact of varying blur thresholds is investigated in Fig.~\ref{fig:threshold_ratio_compare}, where we observe that higher thresholds result in a higher qualified frame ratio, but setting the threshold too high can lead to unnecessary rejection of valid frames. After testing, we set the threshold to 
$T_{blur}=150$, which improves the qualified frame ratio by approximately 15\% for all sets.

We then integrate this filtering method into our object detection pipeline. Blurred frames are excluded and replaced with the last qualified frame, and the detection confidence is smoothed using a moving average filter (MAF). As shown in the subplot of Fig.~\ref{fig:threshold_ratio_compare}, this integration leads to an overall 25\% increase in the qualified frame rate, demonstrating the effectiveness of our motion-blurred frame filtering method.

\subsection{Evaluation on Simulation Environments}

\begin{figure}[t]
    \centering
    \includegraphics[width=1.0\linewidth]{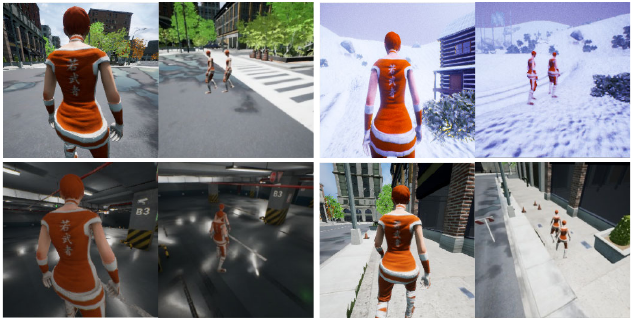}
    \caption{\textbf{Simulation evaluation.} We conduct extensive experiments in Gym-Unreal with four scenes: UrbanCity, SnowVillage, ParkingLot, and Urban Road.}
    \label{fig:sim_demo}
    \vspace{-3mm}
\end{figure}

\noindent\textbf{Benchmark and Evaluation Metric.}
In our evaluation, we follow the setup from previous works~\cite{zhong2023rspt} under the Gym-Unreal benchmark~\cite{gymunrealcv2017} (as shown in Fig.~\ref{fig:sim_demo}), where the maximum episode length is 500 steps. The visible region of the tracker is defined as a 90-degree fan-shaped sector with a radius of 750 cm. 
We evaluate performance using two metrics: 1) Episode Length (EL), which measures the average duration of episodes over 100 trials, reflecting the tracker’s long-term performance; 2) Success Rate (SR), which calculates the percentage of successful episodes across 100 trials, indicating the model's robustness.

\noindent\textbf{Performance Comparisons.}
As shown in Table~\ref{tab:results_baseline}, our LOVON outperforms several baseline approaches, achieving a perfect SR of 1.00 across most environments, including ParkingLot, UrbanCity and SnowVillage. Compared to EVT \cite{zhong2024empowering}, LOVON demonstrates superior tracking performance, e.g., 500\thinspace/\thinspace1.00 vs. 484\thinspace/\thinspace0.92 in ParkingLot. Even when compared to the state-of-the-art TrackVLA~\cite{wang2025trackvla}, which achieves 1.00 SR but requires 360 hours of training, LOVON stands out with an efficient training time of just 1.5 hours, offering both high accuracy and significant efficiency.

\begin{table*}\tiny
\renewcommand{\arraystretch}{1.15}
  \centering
  \caption{\small\textbf{Quantitative results compared with baselines in Gym-Unreal environments.} The numbers of each cell represent the Average Episode Length (EL) and Success Rate (SR). The best results are in bold, where our LOVON achieves superior results compared with previous baselines. $^{*}$DiMP uses a pretrained video tracker which does not need additional training time.}
  \label{tab:results_baseline}
  \resizebox{.9\linewidth}{!}{
  \begin{tabular}{lcccccc}
    \hline\hline
    \textbf{Methods}  & \textbf{Training time} & \textbf{ParkingLot} & \textbf{UrbanCity} & \textbf{UrbanRoad} & \textbf{SnowVillage} & \textbf{Mean} \\
    \hline
    DiMP \cite{bhat2019learning} & 0 hours$^{*}$ & 327\thinspace/\thinspace0.48 & 401\thinspace/\thinspace0.66 & 308\thinspace/\thinspace0.33 & 301\thinspace/\thinspace0.43 & 334.25\thinspace/\thinspace0.48 \\
    SARL \cite{luo2019end} & 24 hours & 301\thinspace/\thinspace0.22 & 471\thinspace/\thinspace0.86 & 378\thinspace/\thinspace0.48 & 318\thinspace/\thinspace0.31 & 367.00\thinspace/\thinspace0.47 \\
    AD-VAT \cite{zhong2019ad} & 12 hours & 302\thinspace/\thinspace0.20 & 484\thinspace/\thinspace0.88 & 429\thinspace/\thinspace0.60 & 364\thinspace/\thinspace0.44 & 394.75\thinspace/\thinspace0.53 \\
    AD-VAT+ \cite{zhong2019ad1} & 24 hours & 439\thinspace/\thinspace0.60 & 497\thinspace/\thinspace0.94 & 471\thinspace/\thinspace0.80 & 365\thinspace/\thinspace0.44 & 443.00\thinspace/\thinspace0.70 \\
    TS \cite{zhong2021towards} & 26 hours & 472\thinspace/\thinspace0.89 & 496\thinspace/\thinspace0.94 & 480\thinspace/\thinspace0.84 & 424\thinspace/\thinspace0.63 & 468.00\thinspace/\thinspace0.83 \\
    RSPT \cite{zhong2023rspt} & 12 hours & 480\thinspace/\thinspace0.80 & \textbf{500}\thinspace/\thinspace\textbf{1.00} & \textbf{500}\thinspace/\thinspace\textbf{1.00} & 410\thinspace/\thinspace0.80 & 472.50\thinspace/\thinspace0.90 \\
    EVT \cite{zhong2024empowering} & 1 hours & 484\thinspace/\thinspace0.92 & \textbf{500}\thinspace/\thinspace\textbf{1.00} & 496\thinspace/\thinspace0.96 & 471\thinspace/\thinspace0.87 & 487.75\thinspace/\thinspace0.94 \\
    TrackVLA \cite{wang2025trackvla}& 360 hours & \textbf{500}\thinspace/\thinspace\textbf{1.00} & \textbf{500}\thinspace/\thinspace\textbf{1.00} & \textbf{500}\thinspace/\thinspace\textbf{1.00} & \textbf{500}\thinspace/\thinspace\textbf{1.00} & \textbf{500.00}\thinspace/\thinspace\textbf{1.00} \\
    \hline
    \textbf{LOVON (Ours)} & \textbf{1.5 hours} & \textbf{500}\thinspace/\thinspace\textbf{1.00} & \textbf{500}\thinspace/\thinspace\textbf{1.00} & 499\thinspace/\thinspace0.99 & \textbf{500}\thinspace/\thinspace\textbf{1.00} & 499.75\thinspace/\thinspace\textbf{1.00} \\
    \hline\hline
  \end{tabular}
  }
  \vspace{-3mm}
\end{table*}

\subsection{Evaluation on Real-World Experiments}
\begin{figure}[t]
    \centering
    \includegraphics[width=1.0\linewidth]{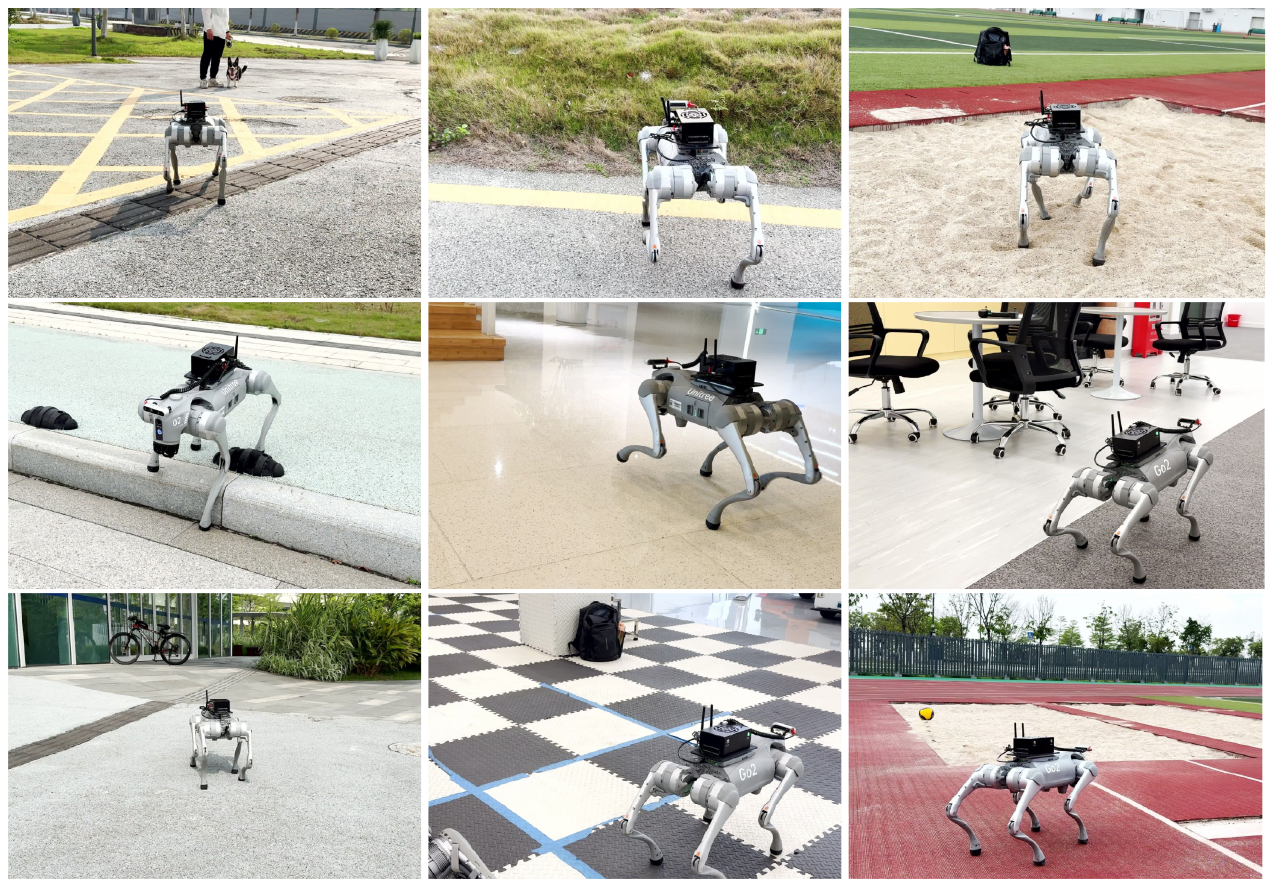}
    \caption{\textbf{Environment adaptation.} LOVON excels in open-world object navigation, effectively adapting to a wide range of objects and environments.}
    \label{fig:multi_envs}
    \vspace{-3mm}
\end{figure}

\begin{figure}[t]
    \centering
    \includegraphics[width=1.0\linewidth]{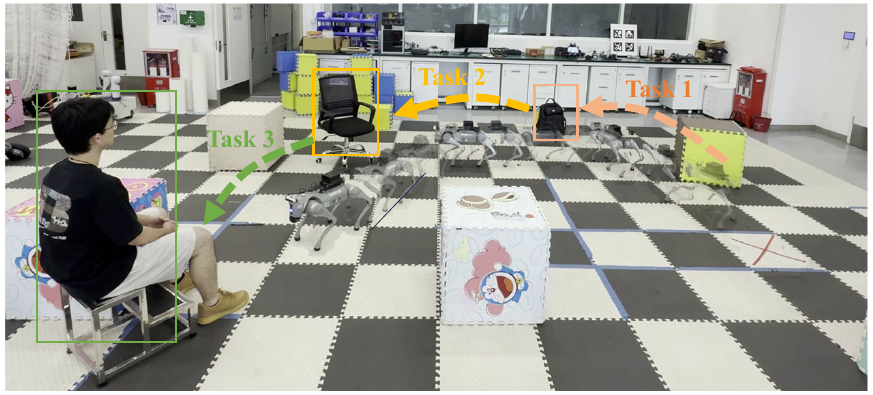}
    \caption{\textbf{Long-horizon tasks with multiple subgoals.} LOVON efficiently handles long-horizon object navigation by coordinating multiple subgoals, ensuring sustained performance over extended tasks.}
    \label{fig:long_horizon}
    \vspace{-3mm}
\end{figure}

In real-world evaluations, LOVON demonstrates exceptional performance in four key areas. (1) First, it excels in \textit{open-world adaptation} (Fig.~\ref{fig:multi_envs}), allowing the robot to handle a wide range of objects commonly encountered in daily life, including large objects like cars, medium-sized ones like people, and small items such as bags. This enables LOVON to seamlessly interact with various objects, regardless of their size or type, in unfamiliar environments.
(2) Second, LOVON achieves \textit{multi-goal tracking} through our LLM planner, enabling long-horizon object navigation. (Fig.~\ref{fig:long_horizon}). This capability allows the robot to efficiently track multiple objects over extended periods, even when the environment becomes more complex.
(3) Third, LOVON excels in \textit{dynamic tracking}, successfully following moving objects in dynamic environments, much like walking a dog. We test this feature on flat roads, spiral stairs, and wild grass, and the robot reliably completes the task in these challenging conditions.
(4) Finally, LOVON is \textit{robust to disturbances}. If the targeted object is displaced or if the robot itself is disturbed (such as being kicked), the robot quickly re-localizes and continues its search. In one test, when we move the chair and kicked the robot, LOVON still enables the robot to approach the chair. Additionally, in a playground with sandy terrain, even when a sports ball is kicked away, the robot completes its task without difficulty.

\subsection{Ablation Study}

\noindent\textbf{Ablation on the Model Parameters.}
We investigate the impact of model size, dataset size $N_{ds}$, motion loss weight $\beta$, and the inclusion of special tokens . The model's performance is evaluated by comparing the standard deviation $\sigma_v$ and the average speed bias $\epsilon_v$ compared to the speed given in the instruction (0.40~m/s). As shown in Table~\ref{tab:model_parameters}, the base-size model shows good performance, while smaller models exhibit higher $\sigma_v$ and $\epsilon_v$, indicating they cannot effectively capture the required information. Larger models generally perform better, but the improvement in speed tracking is marginal despite the model size being much larger. Dataset size affects model stability. Models trained on smaller datasets show lower  $\epsilon_v$ but higher $\sigma_v$, reflecting instability. Larger datasets improve performance but do not significantly impact real-world applicability.
The motion loss weight $\beta$ plays a crucial role. A smaller $\beta$=1 results in poor performance, as the model gives insufficient attention to motion loss. Conversely, a larger $\beta$=20 leads to undervaluation of state loss, causing inaccurate state inference.
Finally, we find that the inclusion of the special token [SEP] is essential. The model trained without this token struggles to follow the given speed, as [SEP] helps differentiate between various input components, especially language.

\begin{table}[t]
\renewcommand{\arraystretch}{1.15}
\caption{\textbf{Ablation study on model parameters.}
We train all the models for 50 epochs and select the model with the lowest validation loss for evaluation.
}
\label{tab:model_parameters}
\centering
\scalebox{0.8}{
\begin{tabular}{l|ccccc}
\hline\hline
\textbf{Model Size} & {$N_{ds}$ (m)} & {$\beta$} & {[SEP]} & {$\sigma_v(\times10^{-4})$}~$\downarrow$ & {$\epsilon_v(\times10^{-2})$}~$\downarrow$ \\ \hline
LOVON-Small (0.47M)               & 0.4  & 10 & \checkmark & 3.74                    & 5.74                     \\ \hline
LOVON-Large (25.51M)               &    0.4                   & 10                    &      \checkmark              & 0.43                    & 0.23                     \\ \hline
\multirow{5}{*}{LOVON-Base (3.30M) } & 0.2                  &  10                  &       \checkmark             & 3.73                    & 0.43                     \\ 
                    & 0.8                  &   10                 &      \checkmark              & \multicolumn{1}{c}{0.40}   & \multicolumn{1}{c}{1.49}    \\  
                    & 0.4  & 1                  &   \checkmark                 & 4.59                    & 7.88                     \\   
                    &              0.4         & 20                 &     \checkmark               & Fail                       & Fail                        \\ 
                    &          0.4             & 10                 & $\times$                  & 1.02                    & 29.5                     \\ \hline
\textbf{LOVON-Base (Best)}              & \textbf{0.4}                   & \textbf{10}                 & \checkmark            & \textbf{0.38}                    & \textbf{1.80}                     \\ 
\hline\hline
\end{tabular}}
\vspace{-3mm}
\end{table}

\noindent\textbf{Ablation on the Filter Method and the Number of States.}
We evaluate the impact of the number of searching states and the frame-filtering technique on the efficiency of object navigation when the target is lost. Three system configurations are tested: Case 1, with three states and no frame filtering; Case 2, with four states but no frame filtering; and Case 3, with four states and frame filtering. Experiments are conducted on the Go2 robot with three objects (backpack, chair, and person) at distances of 4m and 6m.

As shown in Table~\ref{tab:efficiency_time}, LOVON excels in seeking the person, which is easily detected, while the backpack, which is harder to track, requires more effort. In Case 1, the robot performs inefficiently, often losing the object and requiring significant time due to motion blur. Adding a mission state in Case 2 improves efficiency by enabling the robot to search in both directions, though object loss and shaking still occur, leading to a higher number of searching circles.
In Case 3, with four states and the frame-filtering technique, LOVON achieves optimal efficiency. It reduces $N_s$ to 1 for both the backpack and chair, matching the performance with the person. Furthermore, the time to reach the target is reduced by 5 and 2 times for the backpack and chair, respectively, compared to Case 1, approaching the performance observed for the person.

\begin{table}[t]
    \renewcommand{\arraystretch}{1.15}
    \caption{\textbf{Ablation study on proposed methods.} $N_s$: Number of Searching; $T_s$: Search Time (s)}
    \label{tab:efficiency_time}
    \centering
    \scalebox{.9}{
    \begin{tabular}{c|cc|cc|cc}
        \hline\hline
        \multirow{2}{*}{\textbf{Object}} & \multirow{2}{*}{\textbf{States}} & \multirow{2}{*}{\textbf{Filter}} & \multicolumn{2}{c|}{4~m} & \multicolumn{2}{c}{6~m} \\
         & &  & $N_s$~$\downarrow$ & $T_s$~$\downarrow$ & $N_s$~$\downarrow$ & $T_s$~$\downarrow$ \\
        \hline
        \multirow{3}{*}{\shortstack{Backpack}} & 3 & $\times$ &   4.05 & 100.00 & 5.95 & 178.00 \\
        & 4 &$\times$ &  1.60 & 45.91 & 2.25 & 83.09 \\
        & 4 & \checkmark & 1.00 & 22.25 & 1.00 & 32.57 \\
        \hline
        \multirow{3}{*}{\shortstack{Chair}} & 
        3 & $\times$  & 2.05 & 43.50 & 4.25 & 110.00 \\
         & 4 & $\times$  & 1.05 & 26.29 & 1.35 & 45.34 \\
         & 4 & \checkmark  & 1.00 & 21.73 & 1.00 & 30.55\\
        \hline
        \multirow{3}{*}{\shortstack{Person}} & 
        3 & $\times$ &  1.00 & 19.80 & 1.00 & 28.00 \\
         & 4 & $\times$  & 1.00 & 17.66 & 1.00 & 26.24 \\
         & 4& \checkmark & 1.00 & 17.47 & 1.00 & 26.51 \\
        \hline\hline
    \end{tabular}}
    \vspace{-3mm}
\end{table}

\section{Conclusion}
In conclusion, we introduce LOVON, a model designed to tackle long-horizon tasks, adapt to unstructured environments, and achieve state-of-the-art performance in tracking. By incorporating the Laplacian-Variance-based frame filter and the smoothing average confidence filter, we significantly enhance the model's performance in real-world applications. Extensive simulations and real-world experiments across a wide range of complex, open-world scenarios, spanning both indoor and outdoor environments, demonstrate the robustness and effectiveness of LOVON. Moving forward, we aim to refine LOVON’s architecture, enhancing its integration with cutting-edge visual language models to further improve its capabilities in embodied intelligent navigation tasks.

\bibliographystyle{IEEEtran}
\bibliography{mybibs}

\end{document}